\documentclass{article}

\PassOptionsToPackage{numbers,sort&compress}{natbib}
\PassOptionsToPackage{dvipsnames,table}{xcolor}

\makeatletter
\providecommand{\@trackname}{NeurIPS 2026 Workshop}
\makeatother

\usepackage[final]{neurips_2026}

\usepackage{times}
\usepackage{microtype}
\usepackage{graphicx}
\usepackage{booktabs}
\usepackage{amsmath,amssymb}
\usepackage{multirow}
\usepackage{enumitem}
\usepackage{hyperref}
\usepackage{mlpaper_visual_style}
\usepackage{caption}
\usepackage{subcaption}
\usepackage{wrapfig}
\usepackage{float}
\usepackage{tabularx}

\newcommand{\dmargin}{\Delta M}
\newcommand{\model}{OLMo-3-7B-Instruct}

\title{When Goals Leak Across Relationships: How Cached Goal Representations Shape Cross-Client Decisions}

\author{
  Saloni Gupta \\
  University of Maryland
  \And
  S\'ebastien Krier \\
  \And
  Glenn Matlin \\
  Georgia Institute of Technology
}
\begin{document}
\maketitle
\begin{abstract}
As LLM agents take on more autonomous roles, they may interact with multiple clients and counterparties whose goals and incentives conflict. If an agent maintains shared inference state across these interactions, information associated with one relationship may influence decisions made for another. We test OLMo-3-7B-Instruct in a controlled two-client environment with 20 market scenarios, where each client is assigned an opposing objective. With shared conversational memory, the model deviates from the current client’s objective in 22\% of eligible decisions. We then investigate how these goals are represented in the KV cache. Linear probes show that client-goal ownership becomes highly recoverable in middle layers and is concentrated in particular attention heads. We use these results to identify where to intervene in the cache and test whether these representations affect the model's decisions. Ablating the competitor’s cached goal representations at these locations shifts the model toward the current client. The effect is larger for heads identified by probing than for randomly selected heads, and larger when ablating goal tokens than matched non-goal tokens from the same client. Our results show that client-specific goal representations persist in the KV cache and can causally influence decisions made for other clients, highlighting the importance of relationship-specific state isolation in persistent LLM agents.
\end{abstract}

\section{Introduction}

An LLM agent acting on behalf of a user may navigate multiple social relationships, interacting with other agents and parties whose interests do not necessarily align with those of the user it represents. Such an agent may need to retain context across relationships while ensuring that its decisions remain aligned with the objectives of the client it currently serves. For example, an agent advising clients with different investment objectives should preserve relevant market information without allowing one client's preferences to influence recommendations for another. This raises a fundamental question of state isolation: what information should remain shared across interactions, and what should remain specific to each relationship?

When interactions share a persistent KV cache, autoregressive inference does not itself enforce relationship-specific boundaries within that state. Once conflicting client goals enter the same KV cache, both remain available to subsequent attention. This creates a potential memory-isolation problem: information associated with one client remains available when the model later makes decisions for another.

\begin{figure}[t]
    \centering
    \includegraphics[width=\linewidth]{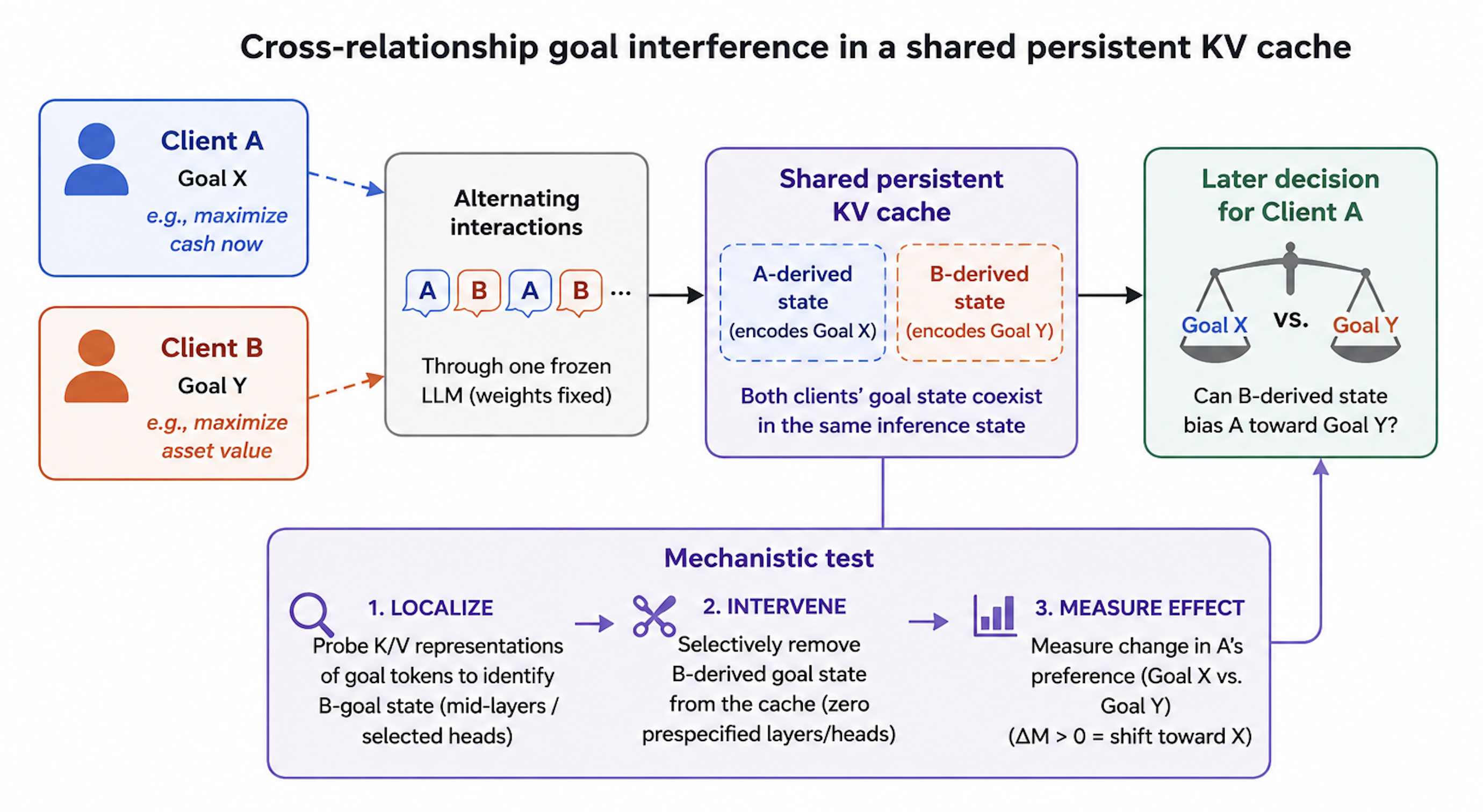}
    \caption{\textbf{Study overview.} Conflicting client goals coexist in a shared persistent KV cache. We localize relationship-conditioned goal state and selectively ablate it to test its effect on later decisions.}
    \label{fig:overview}
\end{figure}

We test this failure mode in a controlled two-client market benchmark using a frozen \model{} with persistent \texttt{past\_key\_values}. Each scenario assigns conflicting goals to Clients A and B, alternates their turns through one shared cache, and later queries the model separately for each client. Isolated controls filter out scenarios the model cannot solve for the queried client on its own. We then trace the interference mechanistically. Linear probes localize goal ownership in the contextualized K/V representation of the natural-language goal tokens; the resulting layer/head locations are frozen before causal evaluation; and selective cache ablations test whether those representations influence the final choice (overview in Figure~\ref{fig:overview}).

Our results support three claims:
\begin{itemize}
\item \textbf{Client goals interfere across relationships.} Under persistent KV reuse, the model deviates from the current client's objective in 22.0\% of eligible decisions. A recompute-from-text baseline also shows interference, so persistent caching is not required for the behavioral effect.

\item \textbf{Client-goal ownership is recoverable from the KV cache.} Linear probing shows that information distinguishing which client a goal belongs to emerges strongly in the middle layers and is concentrated in particular attention heads.

\item \textbf{These representations causally influence later decisions.} Ablating a competitor's cached goal representations shifts behavior toward the current client, while ablating the current client's representations shifts behavior in the opposite direction. Probe-selected heads also have a larger effect than random heads, and goal tokens have a larger effect than matched non-goal context.

\end{itemize}

Client identity and goals are explicit experimental inputs so we therefore do not interpret the probes as evidence that the model discovers latent social relationships. The result is narrower: explicit relationship--goal assignments become structured in persistent KV state, and that state can transmit interference across client contexts.

\section{Related Work}

\paragraph{Agent memory and multi-principal interaction.}
LLM agents increasingly maintain information across interactions through retrieval stores \cite{packer2023memgpt,park2023generative} and persistent KV caches \cite{kwon2023efficient}. Prior work shows that models can become confused or distracted by competing context \cite{liu2024lost,shi2023large}, and that conflicting objectives can interfere with agent behavior \cite{el2025moloch}. This becomes especially relevant when the same agent serves multiple clients: information useful for maintaining one relationship may conflict with the goals of another. We study this setting by asking whether client goals leak across interactions when they share the same conversational memory.

\paragraph{KV-cache persistence and latent sharing.}
Recent work shows that KV state can carry useful information across interactions. KVCOMM directly shares cached state between agents and finds that this information can be reused without reducing task performance \cite{ye2025kvcomm}, while LatentMAS similarly communicates through shared latent working memory \cite{zou2026latent}. The same persistence creates privacy risks. Shadow in the Cache shows that private information can be recovered from KV representations \cite{luo2025shadow}, while CachePrune identifies which cache segments can be safely reused based on their sensitivity and contextual dependence \cite{wu2026cacheprune}. Other work finds that modifying transported KV state can change another agent's downstream behavior even when its visible message remains unchanged \cite{wang2026outofsight,brito2026latent}. These results establish that KV caches retain recoverable, behaviorally relevant information. We ask whether this also happens naturally across relationships, without an attacker modifying or extracting the cache.

\paragraph{Mechanistic probing and causal interventions.}
Linear probes provide a simple way to test what information is recoverable from internal representations \cite{alain2016understanding,belinkov2022probing,hewitt2019designing}, but decodability alone does not show that the model uses that information. Causal interventions such as activation patching and mediation analysis address this by directly manipulating internal representations and measuring changes in behavior \cite{vig2020causal,geiger2021inducing,meng2022locating,wang2023interpretability,zhang2024towards}. We use this combination to first locate client-goal information in the KV cache and then test whether removing that state changes which client's objective the model follows.

Related work on entity binding and retrieval mechanisms \cite{feng2024binding,gurarieh2025mixing} further motivates distinguishing ownership representations from their downstream use. Our experiments establish causal localization of relationship-conditioned goal state, without identifying a complete binding or retrieval mechanism.

\section{Experimental Setup}

\paragraph{Scenarios.}
We construct 20 market-style goal-conflict families (ex: liquidity vs.\ growth, security vs.\ convenience). Each family has a \textsc{base} assignment ($A\!\to\!X,B\!\to\!Y$) and a \textsc{swap} assignment ($A\!\to\!Y,B\!\to\!X$), yielding 40 configurations. Base/swap ensures that the same semantic goal occurs under both ownership labels. For mechanistic experiments, each configuration is run under eight randomized six-turn A/B orders, yielding 320 episodes and 1,920 client turns.

Table 1 provides a schematic illustration of a goal-conflict scenario (Liquidity vs. Growth). In this example, Client A requests high liquidity for near-term obligations while Client B requests capital appreciation via illiquid commitments. In the swap assignment, client goals are inverted. Dialogue turns present alternating client instructions, and evaluation queries present three discrete options (target-aligned, opposite-aligned, and a balanced hedge) without explicitly mentioning competitor goals.

\begin{table}[H]
\centering
\small
\caption{\textbf{Illustrative goal-conflict scenario (Liquidity vs.\ Growth).} This schematic example demonstrates the opposing client objectives and decision options; experimental scenarios use the fixed prompt templates described in Section~3.}
\label{tab:example_scenario}
\begin{tabularx}{\linewidth}{lX}
\toprule
\textbf{Element} & \textbf{Content} \\
\midrule
Client A Goal & Maintain strict liquidity reserves; avoid capital lockups exceeding 30 days. \\
Client B Goal & Maximize compound annualized growth; prioritize high-yield long-term allocations. \\
\midrule
Dialogue Turn 1 & \texttt{[CLIENT A]}: Our board requires immediate cash reserves for upcoming payroll. \\
Dialogue Turn 2 & \texttt{[CLIENT B]}: We seek multi-year fund deployment to capture market premium. \\
\midrule
Evaluation Query & \texttt{[CLIENT A]}: Please allocate our uncommitted portfolio funds across available instruments. \\
\midrule
Target Option & Allocate 100\% to 14-day treasury bills and overnight liquidity facilities. \\
Opposite Option & Commit 100\% to private equity venture funds with 5-year lockup horizons. \\
Mixed Option & Split 50\% into overnight repo facilities and 50\% into illiquid 3-year debt notes. \\
\bottomrule
\end{tabularx}
\end{table}

\paragraph{Persistent memory.}
We study frozen \model{} \cite{groeneveld2024olmo} with deterministic decoding. Previous tokens are processed once; their \texttt{past\_key\_values} are retained and reused as new client turns are appended. Thus, old goal-token K/V entries persist across subsequent A/B switches rather than being reconstructed from the full transcript.

\paragraph{Behavioral decision points.}
For each target client, the model chooses among a target-aligned option, an opposite-client-aligned option, and a mixed option. We report target/opposite/mixed choice rates. For causal interventions, our continuous outcome is
\begin{equation}
    M=\log P(\text{target})-\log P(\text{opposite}),\qquad
    \dmargin=M_{\text{intervention}}-M_{\text{baseline}}.
\end{equation}
Positive $\dmargin$ means the intervention shifts preference toward the current client's objective relative to the competitor's objective. Option probabilities $P(\text{option})$ are computed from model token log-probabilities conditioned on the prompt and active KV state. For behavioral results, the mixed option is counted as a deviation from the current client's target, while $M$ directly compares preference for the current client's objective against the competitor's. Ablations are performed directly on the stored post-RoPE key and value tensors in the KV cache.

\section{Behavioral Interference Persists Under KV Reuse}
Across the 150 eligible persistent-KV decision points, drawn from 20 scenario families, the model deviates from the current client's objective in 22.0\% of decisions (95\% CI $[16.1, 29.3]$\%). In 11.3\% of cases, it selects the opposite client's objective outright; the remaining deviations select the mixed option. Interference occurs for both clients and at both evaluation points (Table~\ref{tab:behavior}). Eligibility is determined by isolated-client controls, so we count failures only when the model successfully follows the target objective without the competing client's context.

\begin{table}[H]
\centering
\small
\caption{\textbf{Persistent-KV behavioral decision points.}
Eligible cases require the corresponding isolated-client control to follow the target objective across 20 scenario families. Target deviation comprises mixed and opposite-client choices; brackets report Wilson 95\% confidence intervals.}
\label{tab:behavior}
\begin{tabular}{lccccc}
\toprule
Decision Point & Target & Eligible & \textbf{\safeconcept{Target}} (\%) & \textbf{\evalconcept{Mixed}} (\%) & \textbf{\riskconcept{Opposite}} (\%) \\
\midrule
Early A & A & 37 & \improved{78.4} [62.8, 88.6] & \cellcolor{MLPurple!8}10.8 [4.3, 24.7] & \regressed{10.8} [4.3, 24.7] \\
Early B & B & 36 & \improved{80.6} [65.0, 90.2] & \cellcolor{MLPurple!8}5.6 [1.5, 18.1] & \regressed{13.9} [6.1, 28.7] \\
Late A  & A & 39 & \improved{79.5} [64.5, 89.2] & \cellcolor{MLPurple!8}12.8 [5.6, 26.7] & \regressed{7.7} [2.7, 20.3] \\
Late B  & B & 38 & \improved{73.7} [58.0, 85.0] & \cellcolor{MLPurple!8}13.2 [5.8, 27.3] & \regressed{13.2} [5.8, 27.3] \\
\bottomrule
\end{tabular}
\end{table}

\paragraph{Recompute baseline.}
Cross-client interference also occurs when the accumulated conversation is reprocessed from text rather than retained through a persistent KV cache. Opposite-client choice rates are 2.6\% for both clients at early checkpoints and 7.7\% and 15.8\% for A and B at late checkpoints, respectively. The two implementations use different inference paths: recomputation applies the model's chat template to the accumulated transcript, whereas persistent execution incrementally appends tokens to a retained cache. We have not established token-level equivalence between these paths, so differences in their observed rates cannot be attributed solely to cache persistence. We use persistent KV reuse to study the internal state available across client interactions, rather than to claim that persistence uniquely causes interference.

\section{Goal Ownership Is Localized in Contextualized KV State}\label{sec:probing}

Having established behavioral leakage under persistent KV-cache reuse, we next investigate whether relationship-conditioned information is explicitly represented within the transformer's KV cache. We know which goal tokens correspond to Client A or Client B from our labeled prompts. The goal is then to use a linear goal-ownership probe to test, at each layer and attention head, whether the contextualized K/V representations of these goal tokens can distinguish whether the goal belongs to Client A or Client B. This allows us to understand where relationship-specific goal information is most recoverable within the KV cache.

For each client turn, prompt construction records the exact token span corresponding to the natural-language goal statement. We train linear logistic regression probes with L2 regularization ($C = 1.0$) on concatenated, span mean-pooled K and V features extracted from these goal-token spans $S$. The client label tokens (e.g., \texttt{[CLIENT A/B]}) are excluded from pooling so that the probe reads the contextualized goal representations rather than simply detecting the client markers, although the goal tokens can attend to earlier ownership context. We fit the probes to predict whether each goal belongs to Client A or Client B. Evaluation uses 5-fold cross-validation grouped by scenario family, ensuring that all base, swap, and order variants of a family are held out together. A 100-shuffle permutation baseline, which randomly shuffles client ownership labels across families, defines the empirical null distribution.

\begin{figure}[!t]
\centering
\includegraphics[width=.88\linewidth]{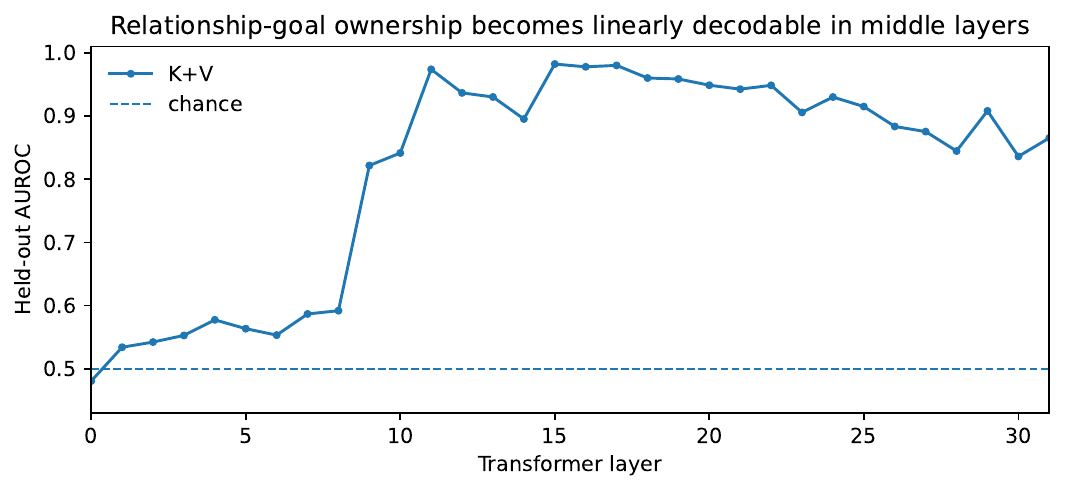}
\caption{\textbf{Layerwise goal-ownership decodability.} Held-out AUROC for concatenated K+V features pooled over tracked goal-token spans. Ownership is near chance at layer 0 (0.481), rises sharply around layers 9--11, and peaks at layer 15 (0.982; null 95th percentile 0.521; $N=1{,}920$; 0/100 permutation shuffles exceeded the observed AUROC, empirical $p < 0.01$).}
\label{fig:probe}
\end{figure}

Figure~\ref{fig:probe} shows where this goal-ownership information becomes recoverable. K+V ownership decodability is near chance at layer 0 (AUROC 0.481), rises sharply through layers 9--11 (0.822, 0.841, 0.974), and peaks at layer 15 (0.982). Looking at individual attention heads further identifies specific locations where goal ownership is highly decodable (e.g., K layer 15 head 16: AUROC 0.993). Importantly, these results tell us where goal ownership is linearly accessible within the KV cache, but do not yet show that the model actually uses this information to make its decisions.

\paragraph{Auxiliary probe controls.}
As a sanity check, we also evaluate auxiliary probes for explicit client identity and relationship history. A current-client identity probe confirms that client markers are already decodable at layer~0, where the prompt labels are directly visible. Goal ownership behaves differently: it is near chance at layer~0 (0.481 AUROC) and only becomes strongly decodable in the middle layers, peaking at 0.982 at layer~15. A relationship-history probe predicts whether the competing client has appeared previously, evaluated on a balanced dataset within turn number $\times$ current client. These auxiliary probes are not used to support our primary mechanistic claim, but help distinguish mid-layer goal ownership from simply extracting explicit client markers.

\section{Causal Ablation of Relationship-Conditioned Cache State}\label{sec:causal}
The previous experiment identified where relationship-conditioned goal information was most recoverable in the KV cache. We now test whether those same representations actually affect the model's decisions.

At each six-turn decision point, we know the exact token positions of the competing client's three previous goal spans from our labeled prompts. The KV cache stores K and V representations for these previous tokens at each layer and attention head. We can therefore modify the cached representations of the competitor's earlier goal tokens before running the checkpoint decision. We zero selected K or V vectors at these positions and rerun the same checkpoint, measuring how the model's preference between the current client's target and the competitor's objective changes. We score A and B separately (640 evaluations per intervention) and restore the original cache before each intervention. K and V interventions are run separately, without changing model weights or cache dimensions.

\subsection{Layer localization}
\begin{figure}[!t]
  \centering
  \includegraphics[width=.88\linewidth]{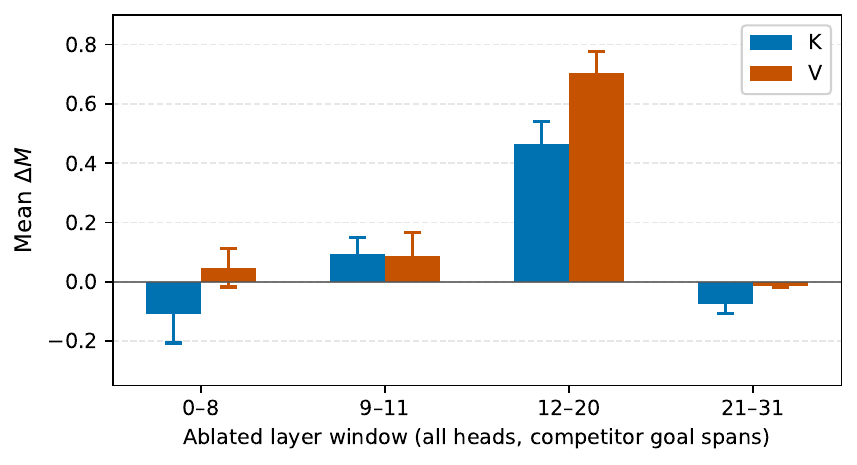}
  \caption{\textbf{Layer-window causal localization.} Mean change in target-vs-opposite log-probability margin after zeroing competitor-goal \modelconcept{K} (blue) or \dataconcept{V} (vermilion) across all heads at the indicated layers. Error bars are 95\% family-clustered bootstrap confidence intervals.}
  \label{fig:causal_layers}
\end{figure}

We first test which layers have the largest effect on the checkpoint decision. We zero the competitor's cached goal state across all 32 heads in prespecified layer windows. The strongest effect occurs across layers 12--20: $\dmargin=+0.464$ for K and $+0.704$ for V (Fig.~\ref{fig:causal_layers}). Removing the competitor's goal state from these layers shifts the model toward the current client's objective.

Zeroing K and V changes the cache in different ways. Zeroing a key vector sets its query-key inner product to zero, giving it an unnormalized attention weight of $\exp(0)=1$. This does not remove the token from attention, but instead redistributes attention mass. Zeroing a value vector prevents that position's value vector from contributing to the attention output when the position is attended to. We find a larger behavioral shift for V-ablation ($\dmargin=+0.704$) than K-ablation ($+0.464$). Layers 9--11 show smaller positive effects, while the early and late layer windows are small or negative. The strongest causal effects are therefore in the same middle layers where we found the highest goal-ownership AUROC.

\subsection{Probe-selected heads beat matched random heads}
We next test whether the heads with the most recoverable goal-ownership information also have a larger effect on the model's decisions. Using the probing results from Section~\ref{sec:probing}, we select the Top-3 heads with the highest goal-ownership AUROC in each layer from layers 9--20, separately for K and V. These were the layers where relationship-specific goal information was most recoverable. We fix the Top3 selection from the probing results before running the causal comparison.

For each checkpoint, we take the competitor's earlier goal-token positions and zero their K or V representations at these Top3 heads. For example, if Client B is the current client, we zero the cached K or V representations of Client A's goal tokens at the selected heads and rerun B's checkpoint decision. We then measure how the target-vs-competitor preference changes relative to the original cache.

We compare this with Random3, where we randomly select three non-Top3 heads from each of the same layers and zero them at the same competitor goal-token positions. Random3 therefore keeps the tokens, layers, number of heads, and type of intervention fixed; the only difference is which heads are selected.

Top3 ablation shifts the model further toward the current client's objective than Random3. For K, Top3 gives $\dmargin=+0.224$ compared with $+0.087$ for Random3. For V, Top3 gives $+0.138$ compared with $+0.030$. The paired Top3-minus-Random3 effects are $+0.137$ for K (95\% CI $[0.090,0.183]$) and $+0.107$ for V ($[0.028,0.180]$), with both confidence intervals excluding zero. Random heads can still affect the decision, but the heads independently identified by our probing experiment have a substantially larger directional effect.

\subsection{Goal state exceeds matched non-goal context}

We also test whether the effect comes specifically from the competitor's goal tokens or from perturbing competitor context more generally. For each decision point, we select the same number of competitor-derived tokens outside the labeled goal spans and run the same interventions. The layers, heads, and number of token positions are held fixed; only the competitor tokens being ablated change.

Ablating the goal spans produces a larger shift toward the current client in all four comparisons. Across layers 12--20, the goal-minus-non-goal difference is $+0.122$ for K (95\% CI $[0.015,0.223]$) and $+0.283$ for V ($[0.187,0.380]$). At the probe-selected Top3 heads, the differences are $+0.111$ for K ($[0.073,0.149]$) and $+0.222$ for V ($[0.148,0.289]$). The effect is therefore larger when we remove the competitor's goal-bearing state than when we remove the same amount of matched non-goal state.

\subsection{Relationship-specific directionality}
Finally, we reverse which client's goal state is ablated. So far, we have removed the competitor's goal state. We now use the same Top3 intervention to zero the \emph{current client's own} goal state. The direction reverses: current-client ablation gives $\dmargin=-0.363$ for K and $-0.511$ for V. The effect is negative across all 20 scenario families for both K and V
(exact two-sided sign test, $p=2^{-19}\approx1.91\times10^{-6}$).

When we remove the competitor's goal state, the model shifts toward the current client. When we instead remove the current client's own goal state, the model shifts away from that client's objective. This rules out the simpler explanation that perturbing the cache itself pushes the model toward the target.

\paragraph{Effects on selected actions.}
Beyond log-probability margins, we examine whether interventions change the model's selected action. Across 640 causal checkpoints, 166 baseline decisions are not target-aligned. Ablating competitor-derived V state in layers 12--20 changes 23 of these decisions (13.9\%) to the target-aligned action, without changing any initially target-aligned decisions to non-target decisions. The corresponding K intervention corrects 13 non-target decisions while changing two target-aligned decisions to non-target. These results establish that the interventions affect categorical decisions, although changes in selected actions are less frequent than shifts in the log-probability margin.

\begin{figure}[!t]
  \centering
  \includegraphics[width=.88\linewidth]{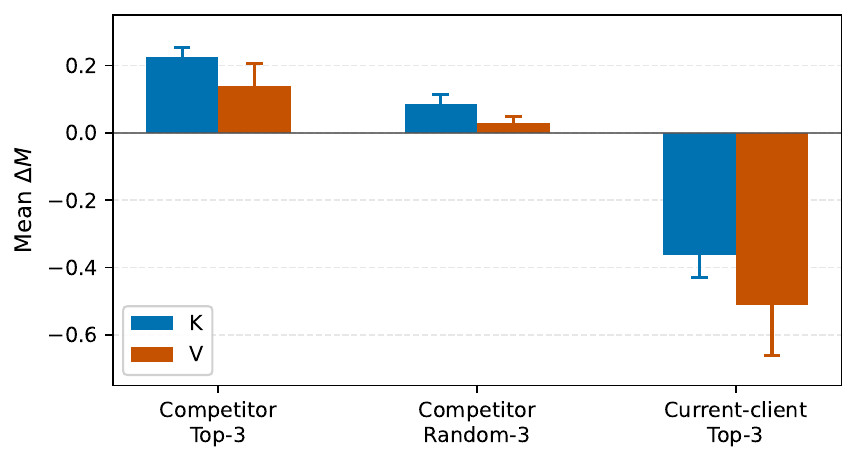}
  \caption{\textbf{Head selection and directionality controls.} Competitor-state ablation at probe-selected Top3 heads shifts preference toward the current client more than matched Random3 ablation. Ablating the current client's own state reverses the direction. Results shown for \modelconcept{K} (blue) and \dataconcept{V} (vermilion). Error bars are 95\% family-clustered bootstrap confidence intervals.}
  \label{fig:heads}
\end{figure}

\section{Discussion and Limitations}
Our experiments support a narrow causal claim: \textbf{explicit relationship--goal assignments become linearly accessible in specific regions of persistent K/V state, and modifying goal-derived state at those locations changes which client's objective the model follows.} Probe-selected heads have a larger effect than random heads, goal tokens matter more than matched non-goal context, and reversing which client's goal is ablated reverses the behavioral effect.

This raises a broader question for persistent agent memory: what information should persist across relationships, and what should remain specific to one client? An agent may need to remember information across interactions without allowing one client's goals to affect decisions for another. Separating relationship-specific state while preserving useful shared memory is a natural next step.

In addition, our experiments are limited to one 7B model and controlled two-client market scenarios with clearly opposing objectives. These scenarios make interference easier to isolate, but do not capture more realistic settings, such as a personal assistant serving different users with conflicting preferences or an agent maintaining separate goals across multiple clients. Client ownership is also explicitly provided rather than inferred. Finally, we use zero-ablation to test causality, which directly modifies cached state but does not represent a naturally occurring change during inference.

These interventions do not distinguish whether cross-client interference arises from incorrect goal/client binding, inappropriate retrieval of competing goal information, or downstream competition between correctly represented objectives. Identifying which mechanism is responsible remains an important direction for future work.

\section{Conclusion}

We find that conflicting client goals can interfere across interactions in an agent with shared persistent state. Goal ownership becomes strongly decodable from middle-layer K/V representations, and selectively modifying this state changes which client's objective the model follows. These results motivate separating relationship-specific state in persistent agents while preserving information that should remain shared across interactions.

\section*{Author Contributions}

\textbf{Saloni Gupta:} Conceptualization, Methodology,
Software, Investigation, Formal Analysis,
Writing -- Original Draft, Writing -- Review \& Editing.

\textbf{Glenn Matlin:} Conceptualization, Methodology.

\textbf{S\'ebastien Krier:} Conceptualization, Supervision.

\bibliographystyle{unsrtnat}
{\emergencystretch=2em\bibliography{references}}

\end{document}